\documentclass[accepted]{uai2023} 

\usepackage[american]{babel}

\usepackage{natbib} 
\renewcommand{\bibsection}{\subsubsection*{References}}
\usepackage{mathtools} 
\usepackage{booktabs} 

\usepackage{hyperref}       
\usepackage{url}            
\usepackage{amsfonts}       
\usepackage{nicefrac}       
\usepackage{microtype}      
\usepackage{enumitem}
\usepackage{wrapfig}
\usepackage{subfig}
\usepackage{graphicx}
\usepackage{multirow}
\usepackage{amsmath}
\usepackage{amssymb}
\usepackage{mathtools}
\usepackage{amsthm}

\newcommand{\bbE}{\mathbb{E}}
\newcommand{\bbR}{\mathbb{R}}

\newcommand{\bw}{\mathbf{w}}

\newcommand{\bx}{\mathbf{x}}
\newcommand{\by}{\mathbf{y}}

\newcommand{\calD}{\mathcal{D}}

\newcommand{\calN}{\mathcal{N}}

\title{Learning To Invert: Simple Adaptive Attacks for Gradient Inversion in Federated Learning}

\author[1 *]{\href{mailto:<rw565@cornell.edu>?Subject=Your UAI 2023 paper}{Ruihan Wu}{}}
\author[1 *]{Xiangyu Chen}
\author[2]{Chuan Guo}
\author[1]{Kilian Q. Weinberger}
\affil[1]{%
    Cornell University\\
    USA
}
\affil[2]{%
    Meta AI\\
    USA
}
\affil[*]{%
equal contribution
  }
  
\begin{document}
\maketitle

\begin{abstract}
Gradient inversion attack enables the recovery of training samples from model gradients in federated learning (FL), and constitutes a serious threat to data privacy. To mitigate this vulnerability, prior work proposed both principled defenses based on differential privacy, as well as heuristic defenses based on gradient compression as countermeasures. These defenses have so far been very effective, in particular those based on gradient compression that allow the model to maintain high accuracy while greatly reducing the effectiveness of attacks.
In this work, we argue that such findings underestimate the privacy risk in FL. As a counterexample, we show that existing defenses can be broken by a simple adaptive attack, where a model trained on auxiliary data is able to invert gradients on both vision and language tasks.
\end{abstract}

\section{Introduction}
\label{sec:intro}

Federated learning (FL; \citep{mcmahan2017communication}) is a popular framework for distributed model training on sensitive user data. Instead of centrally storing the training data, FL operates in a server-client setting where the server hosts the model and has no direct access to clients' data. The clients can apply the model to their private data and send gradient updates back to the server. This learning regime promises data privacy as users share only gradients but never any raw data. 
However, recent work~\citep{zhu2019deep, zhao2020idlg, geiping2020inverting} showed that despite these efforts, the server is still able to recover training data from gradient updates, violating the promise of data privacy in FL. These so-called \emph{gradient inversion attacks} operate by optimizing over the input space to search for samples whose gradient matches that of the observed gradient, and such attacks remain effective even when clients utilize secure aggregation~\citep{bonawitz2016practical} to avoid revealing individual gradients~\citep{yin2021see, jeon2021gradient}.

As countermeasures against these gradient inversion attacks, prior work proposed both principled defenses based on differential privacy~\citep{abadi2016deep}, as well as heuristics that compress the gradient update through gradient pruning~\citep{aji2017sparse} or sign compression~\citep{bernstein2018signsgd}. In particular, gradient compression defenses have so far enjoyed great success, severely hindering the effectiveness of existing optimization-based attacks~\citep{zhu2019deep, jeon2021gradient} while maintaining a similar level of model performance.
As a result, these limitations seemingly diminish the threat of gradient inversion attacks in practical FL applications.

\begin{figure*}[t]
    \centering
    \includegraphics[width=0.88\linewidth]{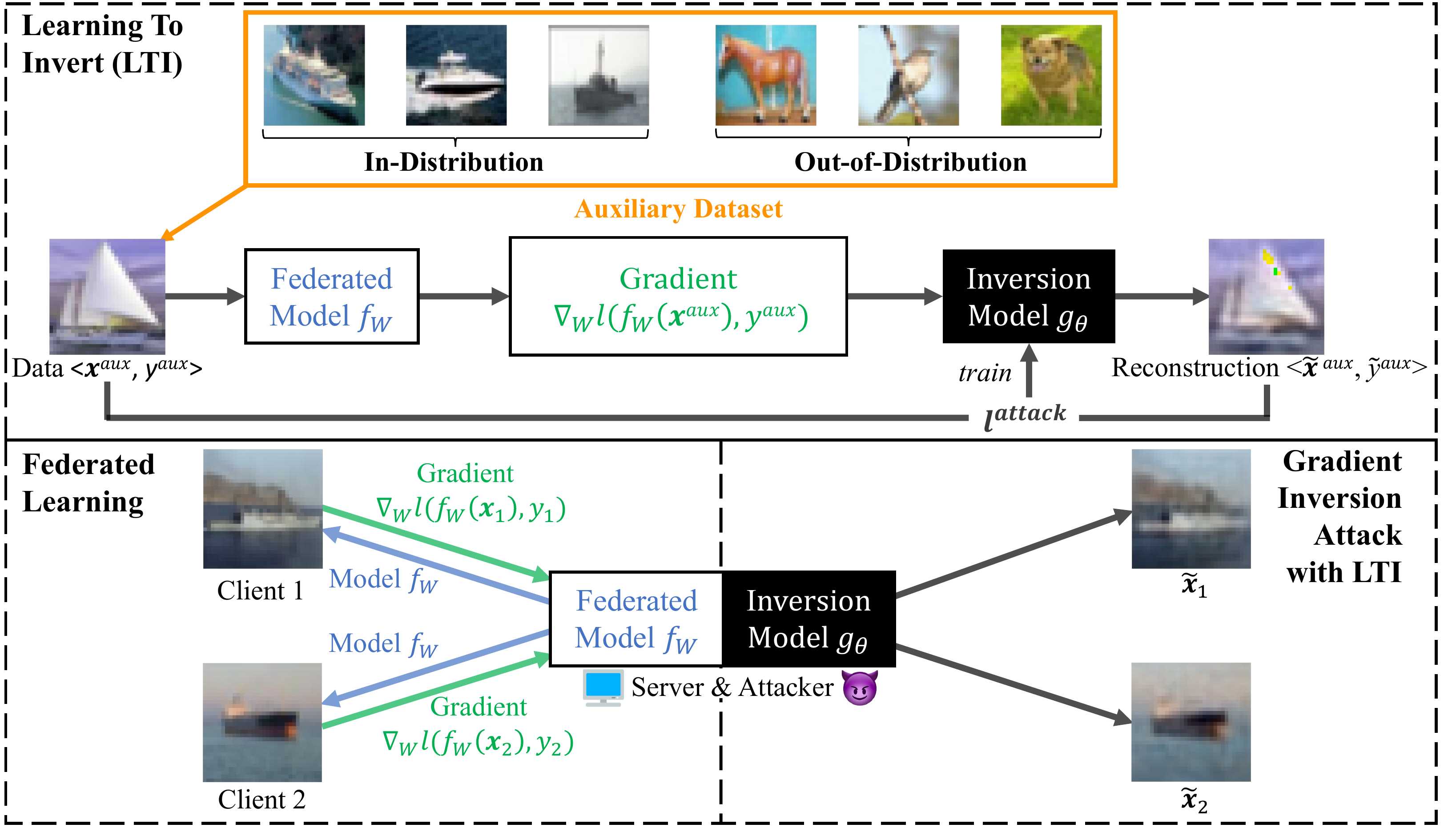}
    \caption{Illustration of federated learning (FL) and gradient inversion methods. The goal of gradient inversion is to recover training data $(\bx, y)$ from the observed gradient $\nabla_{\bw}\ell(f_{\bw}(\bx), y)$. Optimization-based methods (\emph{e.g.}, \citep{zhu2019deep, geiping2020inverting, yin2021see, jeon2021gradient}) directly optimize $(\tilde{\bx}, \tilde{y})$ in search for a sample that produces gradient similar to that of $(\bx, y)$. Our proposed learning-based approach, which we call \emph{Learning to Invert}, instead trains an inversion model $g_\theta$ to reconstruct training samples from their gradient.}
    \label{fig:intro}
\end{figure*}

In this paper, we argue that evaluating defenses on existing optimization-based attacks may provide a false sense of security. To this end, we propose a simple \emph{learning-based} attack---which we call \emph{Learning To Invert} (LTI)---that trains a model to invert gradient updates and recover client samples; see \autoref{fig:intro} for an illustration.
We assume that the adversary (\emph{i.e.}, the server) has access to an \emph{auxiliary dataset} whose distribution is similar to that of the private data. The gradient inversion model trains on samples in the auxiliary dataset, with corresponding gradients provided by the global model. Our attack is highly adaptable to different defense schemes, since applying a defense simply amounts to training data augmentation for the gradient inversion model.

We empirically demonstrate that LTI can successfully circumvent defenses based on gradient perturbation (\emph{i.e.}, using differential privacy; ~\citep{abadi2016deep}), gradient pruning~\citep{aji2017sparse} and sign compression~\citep{bernstein2018signsgd} on both vision and language tasks.
\begin{itemize}[noitemsep, nolistsep, leftmargin=*]
    \item Vision: We evaluate on the CIFAR10~\citep{krizhevsky2009learning} classification dataset for both LeNet and ResNet20. LTI attains recovery accuracy close to that of the best optimization-based method when no defense is applied, and significantly outperforms all prior attacks under defense settings.
    \item NLP: We experiment with both text classification task on CoLA~\citep{warstadt2018neural} and causal language model training on the WikiText~\citep{merity2016pointer} dataset, where LTI attains state-of-the-art performance in all settings, with or without defense.
\end{itemize}
Given the strong empirical performance of LTI and its adaptability to different learning tasks and defense mechanisms, we advocate for its use as a simple baseline for future studies on gradient inversion attacks in FL.
\section{Background}
\label{sec:background}
\paragraph{Federated learning.} The objective of federated learning~\citep{mcmahan2017communication} is to train a machine learning model in a distributed fashion without centralized collection of training data. In detail, let $f_{\bw}$ be the \emph{global federated model} parameterized by $\bw$, and consider a supervised learning setting that optimizes $\bw$ by minimizing a loss function $\ell$ over the training set $\calD_{\rm train}$: $\sum_{(\bx, y)\in \calD_{\rm train}}\ell(f_{\bw}(\bx), y)$. In centralized learning this is typically done by computing a stochastic gradient $\frac{1}{B} \sum_{i=1}^B \nabla_{\bw}\ell(f_{\bw}(\bx_i), y_i)$ over a randomly drawn batch of data $(\bx_1,y_1),\ldots,(\bx_B,y_B)$ and minimizing $\ell$ using stochastic gradient descent.

In FL, instead of centrally collecting $\calD_{\rm train}$, the training set $\calD_{\rm train}$ is distributed across multiple clients and the model $f_{\bw}$ is stored on a central server. At each iteration, the model parameter $\bw$ is transmitted to each client to compute the per-sample gradients $\{ \nabla_\bw \ell(f_{\bw}(\bx_i), y_i) \}_{i=1}^B$ locally over a set of $B$ clients. The server and clients then execute a \emph{federated aggregation} protocol to compute the average gradient for the gradient descent update. A major advantage of FL is data privacy since clients do not need to disclose their data explicitly, but rather only send their gradient $\nabla_\bw \ell(f_{\bw}(\bx_i), y_i)$ to the server. Techniques such as secure aggregation~\citep{bonawitz2016practical} and differential privacy~\citep{dwork2006calibrating, dwork2014algorithmic} can further reduce the risk of privacy leakage from sending this gradient update.

\paragraph{Gradient inversion attack.} Despite the promise of data privacy in FL, recent work showed that the heuristic of sending gradient updates instead of training samples themselves in fact provides a false sense of security. \citet{zhu2019deep} showed in their seminal paper that it is possible for the server to recover the full batch of training samples given aggregated gradients. These \emph{optimization-based} gradient inversion attacks operate by optimizing a set of \emph{dummy data} $\tilde{\bx}_1,\ldots,\tilde{\bx}_B$ and labels $\tilde{y}_1,\ldots,\tilde{y}_B$ to match their gradients to the observed gradients with cost function:
\begin{equation}
    \label{eq:opt_objective}
    \min_{\tilde{\bx}, \tilde{\by}} \left\| \sum_{i=1}^B \nabla_\bw \ell(f_{\bw}(\tilde{\bx}_i), \tilde{y}_i) - \sum_{i=1}^B \nabla_\bw \ell(f_{\bw}(\bx_i), y_i) \right\|_2^2
\end{equation}

For image tasks, since \autoref{eq:opt_objective} is differentiable to $\tilde{\bx}_i$ and $\tilde{y}_i$, and the model parameters $\bw$ are accessible to the server, the server can simply optimize \autoref{eq:opt_objective} using gradient-based search. Doing so yields recovered samples $(\tilde{\bx}_i,\tilde{y}_i)$ that closely resemble actual samples $(\bx_i,y_i)$ in the batch. In practice this approach is highly effective, and follow-up works proposed several optimizations to further improve its recovery accuracy~\citep{geiping2020inverting, yin2021see, jeon2021gradient}.

For language tasks this optimization problem is considerably more complex since the samples $\bx_1,\ldots,\bx_B$ are sequences of discrete tokens, and optimizing \autoref{eq:opt_objective} amounts to solving a discrete optimization problem. To circumvent this difficulty, \citet{zhu2019deep} and \citet{deng2021tag} instead optimize the \emph{token embeddings} to match the observed gradient and then maps the recovered embeddings to their closest tokens in the embedding layer to recover the private text. In contrast, \citet{gupta2022recovering} leveraged the insight that the gradient of the token embedding layer can be used to recover exactly the set of tokens present in the training sample, and used beam search to optimize the ordering of tokens for fluency to recover the private text.

\paragraph{Gradient inversion under the malicious server setting.} The aforementioned gradient inversion attacks operate under the \emph{honest-but-curious} setting where the server faithfully executes the federated learning protocol, but attempts to extract private information from the observed gradients. \citet{fowl2021robbing}, \citet{boenisch2021curious} and \citet{fowl2022decepticons} consider a stronger \emph{malicious server} threat model, which allows the server to transmit arbitrary model parameters $\bw$ to the clients. Under this threat model, it is possible to carefully craft the model parameters such that the training sample can be recovered exactly from its gradient even when the batch size $B$ is large. While this setting is certainly realistic and relevant, our paper operates under the weaker honest-but-curious threat model.
\section{Learning To Invert: Learning-based Gradient Inversion Attacks}
\subsection{Problem Set-Up}
\textbf{Motivation.} The threat of gradient inversion attack has prompted prior work to employ defense mechanisms to mitigate this privacy risk in FL~\citep{zhu2019deep, jeon2021gradient}. Intuitively, such defenses reduce the amount of information contained in the gradient about the training sample by either perturbing the gradient with noise~\citep{abadi2016deep} or compressing them~\citep{aji2017sparse, bernstein2018signsgd}, making recovery much more difficult. However, doing so also reduces the amount of information a sample can provide for training the global model, and hence has a negative impact on the model's performance. This is certainly true for principled defenses based on differential privacy~\citep{dwork2006calibrating} such as gradient perturbation~\citep{abadi2016deep}. However, defenses based on gradient compression seemingly provide a much better privacy-utility trade-off, effectively preventing the attack and reducing communication costs with minor reduction in model performance~\citep{zhu2019deep}. 

The empirical success of existing defenses seemingly diminish the threat of gradient inversion attacks in FL.
However, we argue that optimization-based attacks underestimate the power of the adversary: If the adversary has access to an auxiliary dataset $\calD_{\rm aux}$, they can train a \emph{gradient inversion model} to recover $\calD_{\rm aux}$ from its gradients computed on the global model.
As we will establish later, this greatly empowers the adversary, exposing considerable risks to federate learning. 


\textbf{Threat model.} 
We consider the setting where the adversary is an \textit{honest-but-curious} server, who executes the learning protocol faithfully but aims to extract private training data from the observed gradients. 
Hence, in each FL iteration, the adversary has the knowledge of model weights $\bw$ and aggregated gradients.
Moreover, we assume the adversary has an auxiliary dataset $\calD_{\rm aux}$, which could be in-distribution or a mixture of in-distribution and out-of-distribution data.
This assumption is similar to the setting in \citet{jeon2021gradient}, which assumes a generative model that is trained from the in-distribution data, and is common in the study of other privacy attacks such as membership inference~\citep{shokri2017membership}.

In this paper, we focus on the attack against defense mechanisms ($\mathrm{DM}$) in prior work~\citep{zhu2019deep, jeon2021gradient}. Thus, we assume the adversary receives the aggregated gradients $\sum_{i=1}^B \mathrm{DM}\left[\nabla_{\bw}\ell(f_{\bw}(\bx_i), y_i)\right]$ at one of following $\mathrm{DM}$ settings:
\begin{enumerate}[leftmargin=*,nosep]
    \item \emph{Gradient without defense.} The gradient before the aggregation is the original gradient without any defense. Most previous papers focus on this common setting.
    \item \emph{Sign compression.}~\citep{bernstein2018signsgd} applies a element-wise sign function to gradient before the aggregation, which compresses the gradient to \emph{one bit per dimension}.
    \item \emph{Gradient pruning with pruning rate $\alpha$}~\citep{aji2017sparse} zeroes out the bottom $1-\alpha$ fraction of coordinates of $\nabla_{\bw}\ell(f_{\bw}(\bx), y)$ in terms of absolute value, which effectively compresses the gradient to $(1-\alpha) m$ dimensions, where $m$ denotes the model size. 
    \item \textit{Gradient perturbation with Gaussian standard deviation $\sigma$}~\citep{abadi2016deep} is a differentially private mechanism used commonly for training private models. A Gaussian random vector $\calN(\mathbf{0}, \sigma^2I)$ is added to the gradient, which one can show achieves $\epsilon$-local differential privacy~\citep{kasiviswanathan2011can} with $\epsilon = O(1/\sigma)$.
\end{enumerate}



\subsection{Learning to invert (LTI)}
\label{sec: method}
\textbf{Definition of the learning problem.} 
Having knowledge of the model weights and the defense mechanism $\mathrm{DM}$, the adversary is able to generate the gradient $\mathrm{grad}_{S_B}^{\rm DM}=\sum_{i=1}^B\mathrm{DM}\left[\nabla_{\bw}\ell(f_{\bw}(\bx^{\rm aux}_i), y^{\rm aux}_i)\right]$ for any batch of samples $S_B = \{(\bx^{\rm aux}_1, y^{\rm aux}_1)\cdots (\bx^{\rm aux}_B, y^{\rm aux}_B)\} $ in the auxiliary dataset.
This allows the adversary to learn a \emph{gradient inversion model} $g_{\theta}: \bbR^m\to \bbR^{B\times d}$ ($d$ denotes data dimension), parameterized by $\theta$ , to predict this batch of data point $S_B$ from the aggregated gradient $\mathrm{grad}_{S_B}^{\rm DM}$.
The learning goal is to minimize the reconstruction error $\ell^{attack}$ of $g_{\theta}$ on the auxiliary dataset $\calD^{\rm aux}$:
\begin{equation}
    \label{eq:obj}
    \min_{\theta}\bbE_{S_B\sim\calD^{\rm aux}}\ell^{ attack}\left(g_{\theta}\left(\mathrm{grad}_{S_B}^{\rm DM}\right), S_B \right).
    \end{equation}

We hereby explain the choice of the loss function $\ell^{attack}$ and the inversion model $g_{\theta}$.
Since $g_{\theta}$ needs to reconstruct data in batches, $\ell^{attack}$ should be permutation invariant w.r.t. the $S_B$.
A common solution \citep{zhang2019deep} is to define $\ell^{attack}$ as:
\begin{align}
\label{eq:obj_single}
	&\ell^{attack}\left(g_{\theta}\left(\mathrm{grad}_{S_B}^{\rm DM}\right), S_B \right)\nonumber \\
	&= \min_{\pi}\sum_{i=1}^B \ell^{attack}_{single}
	\left(\left(\mathrm{grad}_{S_B}^{\rm DM}\right)_i, \left(\bx^{\rm aux}_{\pi(i)}, y^{\rm aux}_{\pi(i)}\right)\right),
\end{align}
where the minimization is over all possible permutation $\pi$. 
$\ell^{attack}_{single}$ is the loss function for a single pair of the prediction and target data.
In practice, $\ell^{attack}_{single}$ can be a cross-entropy loss for discrete inputs or a L2 loss for continuous-valued inputs.
As for the choice of the inversion model $g_{\theta}$, we empirically find that a multi-layer perceptron (MLP)~\citep{bishop1995neural} is sufficiently effective for the tasks in our experiments.

\textbf{Comparison to optimization-based attacks.} LTI is superior in its simplicity to optimization-based methods on generalization, for the following two aspects.
Firstly, LTI doesn't explicitly have any terms relevant to data prior. It will learn the data property from the auxiliary dataset. 
However, optimization-based attacks usually manually encode the data prior in their objective functions, e.g. the total variation term in most optimization-based attacks to reconstruct image samples.
Secondly, there's no need for careful adaptation to different defense mechanisms. 
As we know, in optimization-based attacks, for any FL defense mechanism, it is crucial to carefully design a corresponding objective function for gradient matching.
In \autoref{sec:exp}, we will show that our simple approach is surprisingly effective at circumventing existing defenses for both language and vision data.

\textbf{Dimensionality reduction for large models $f_{\bw}$.} One potential problem for LTI is that the gradients $\sum_{i=1}^B\mathrm{DM}\left[\nabla_{\bw}\ell(f_{\bw}(\bx^{\rm aux}_i), y^{\rm aux}_i)\right]$ can be extremely high-dimensional. For example, ResNet20~\citep{he2016deep} for vision tasks has $270K$ parameters and BERT~\citep{devlin-etal-2019-bert} for language tasks have approximately $110M$ trainable parameters. Such high-dimensional input to the model $g_\theta$ can lead to memory issues, as the first layer of the MLP would have $110M \times h$ parameters, where $h$ denotes the size of the first hidden layer.


To address this issue, we use feature hashing~\citep{weinberger2009feature} to reduce the dimensionality of the input gradient.
In feature hashing, each gradient dimension $i\in[m]$ is randomly assigned to one of $k$ bins ($k$ is much smaller than the size of gradient $m$), formalized as $r(i)\in[k]$. 
We then sum up all gradient values in each bin, producing a compressed feature vector of size $k$. 
In other words, we project the aggregated gradient $\sum_{i=1}^B\mathrm{DM}\left[\nabla_{\bw}\ell(f_{\bw}(\bx^{\rm aux}_i), y^{\rm aux}_i)\right]$ to $P\left(\sum_{i=1}^B\mathrm{DM}\left[\nabla_{\bw}\ell(f_{\bw}(\bx^{\rm aux}_i), y^{\rm aux}_i)\right]\right)$ using the random projection matrix $P$ given by:
$$
P \in \{0, 1\}^{k\times m} s.t.~\forall i,~P_{j, i}=0~(\forall j\neq r(i)),~P_{r(i), i}=1.
$$
$P$ in the definition is a sparse matrix with $m$ nonzero element that can be saved in a memory efficient way.
In this way, $g_\theta$ 's the memory footprint can be reduced to a constant independent from the gradient dimension. 


\section{Experiment}
\label{sec:exp}
We evaluate LTI on both vision and language tasks.
The evaluation results demonstrate that it vastly outperforms prior gradient inversion attacks, especially when \emph{gradient defenses are applied}. 
Moreover, we show that LTI is able to perform surprisingly well even when the auxiliary data is out-of-distribution, which makes LTI more applicable in the real scenario\footnote{Our code is released at \url{https://github.com/wrh14/Learning_to_Invert}.}.

\subsection{Evaluation on Vision Task}
\label{sec:exp_vision}
\paragraph{Federate learning tasks.} For evaluating LTI on vision tasks, we experiment with image classification on CIFAR10~\citep{krizhevsky2009learning} and the training loss is the cross-entropy loss. 
The \textit{original test split of CIFAR10} is used for FL training.
For the generalization propose, we test the attacks on two different architectures as the FL model $f_{\bw}$, which are LeNet~\citep{lecun1998gradient} and ResNet20~\citep{he2016deep} with $~15K$ and $~270K$ parameters.

\paragraph{Defense mechanisms set-up.} The adversary will receive the gradient aggregated from $B=1$ or $4$ clients, applied with no defense, sign compression, gradient pruning ($\alpha=0.99$), or Gaussian perturbation ($\sigma=0.1$).

\paragraph{Baselines.}
We compare our method with two gradient inversion attack baseline methods: \emph{Inverting Gradients} (IG; \citet{geiping2020inverting}), a representative optimization-based method with limited data prior, and \emph{Gradient Inversion with Generative Image Prior} (GI-GIP; \citet{jeon2021gradient}), the state-of-the-art optimization-based method that uses a generative model to encode the data prior. We make minor modifications to these attacks to adapt them to various defenses; see appendix for details. The threat model of LTI is most similar to GI-GIP since both use an auxiliary dataset to encode the data prior.

\paragraph{Set-up of LTI.} We introduce the training set-up of LTI. 
\begin{itemize}[leftmargin=*,nosep]
    \item \textit{Auxiliary dataset.} We use the \textit{original train split of CIFAR10} as the auxiliary dataset of the adversary. Notice that under this set-up, the auxiliary dataset is different from the dataset that the FL tasks are trained on, i.e. the one from which the aggregated gradients are computed.
    \item \textit{Inversion model architecture.} Our inversion model $g_{\theta}$ is a three-layer MLP with hidden size 3K or 10K upon the memory limitation.
    The MLP takes the flattened gradient vector as input and outputs a $B\times 3072$-dimensional vector representing the flattened images. 
    Because the size ResNet20 is large, we use feature hashing (see \autoref{sec: method}) to reduce the target model gradient to $50\%$ of its original dimensionality as input to the inversion model.
    \item \textit{Training details.} The training objective $\ell^{attack}_{single}$ in \autoref{eq:obj_single} is the mean squared error (MSE) between the output vector from MLP and the flattened ground truth image. We use the Adam~\citep{kingma2014adam} optimizer for training $g_\theta$. The model is trained for $200$ epochs using training batch size $256$. The initial learning rate is $10^{-4}$ with learning rate drop to $10^{-5}$ after $150$ epochs.
    \item \textit{Computation cost.} Our experiments are conducted using NVIDIA GeForce RTX 2080 GPUs and each training run takes about 1.5 hours.
\end{itemize}

\paragraph{Evaluation methodology.}
We evaluate LTI and the aforementioned baselines on $1,000$ random images from the CIFAR10 test split. 
To measure reconstruction quality, we use three common metrics: 1. \emph{Mean squared error} (MSE) measures the average pixel-wise (squared) distance between the reconstructed image and the ground truth image. 2. \emph{Peak signal-to-noise ratio} (PSNR) measures the ratio between the maximum image pixel value and MSE. 3. \emph{Learned perceptual image patch similarity} (LPIPS) measures distance in the features space of a VGG~\citep{simonyan2014very} model trained on ImageNet. 4. \emph{Structural similarity index measure} (SSIM) measures the perceived change in structural information

\begin{table*}[t!]
\centering
\caption{MSE for baselines (IG and GI-GIP) and our method LTI on CIFAR10. As shown in the table, neither IG nor GIGIP works well when the defense mechanism is applied, while our method has the power to break the privacy protection from the compression and randomness.} 
    \label{tab:res_cv}

    \resizebox{\linewidth}{!}{
    \begin{tabular}{c|c|cccc|cccc}
    \toprule
     \multirow{2}{*}{\textbf{FL model}} & \multirow{2}{*}{\textbf{Methods}} & \multicolumn{4}{c|}{$B=1$}& \multicolumn{4}{c}{$B=4$}\\
     \cmidrule{3-10}
         & & \textbf{None} & \textbf{Sign Comp.} & \textbf{Grad. Prun.} & \textbf{Gauss. Pert.} & \textbf{None} & \textbf{Sign Comp.} & \textbf{Grad. Prun.} & \textbf{Gauss. Pert.}\\ 
        \midrule
        \multirow{3}{*}{LeNet} & IG         & 0.022          & 0.116          & 0.138          & 0.150          & 0.105          & 0.265          & 0.169          & 0.206          \\
 & GI-GIP      & \textbf{0.001} & 0.091          & 0.043          & 0.124          & \textbf{0.009} & 0.082          & 0.180          & 0.157          \\
 & LTI (Ours) & 0.004          & \textbf{0.014} & \textbf{0.029} & \textbf{0.012} & 0.015          & \textbf{0.023} & \textbf{0.031} & \textbf{0.026}\\
 \midrule
\multirow{3}{*}{ResNet20}  &  IG         & 0.120 & 0.154          & 0.171          & 0.133          & 0.125          & 0.272          & 0.195          & 0.123          \\
 & GI-GIP      & 0.062                         & 0.099          & 0.238          & 0.233          & 0.086          & 0.236          & 0.231          & 0.229          \\
 & LTI (Ours) & \textbf{0.018}                & \textbf{0.013} & \textbf{0.023} & \textbf{0.021} & \textbf{0.038} & \textbf{0.035} & \textbf{0.038} & \textbf{0.039}\\
     \bottomrule
    \end{tabular}%
    }
\end{table*}

\begin{figure*}[!t]
    \centering
    \includegraphics[width=\linewidth]{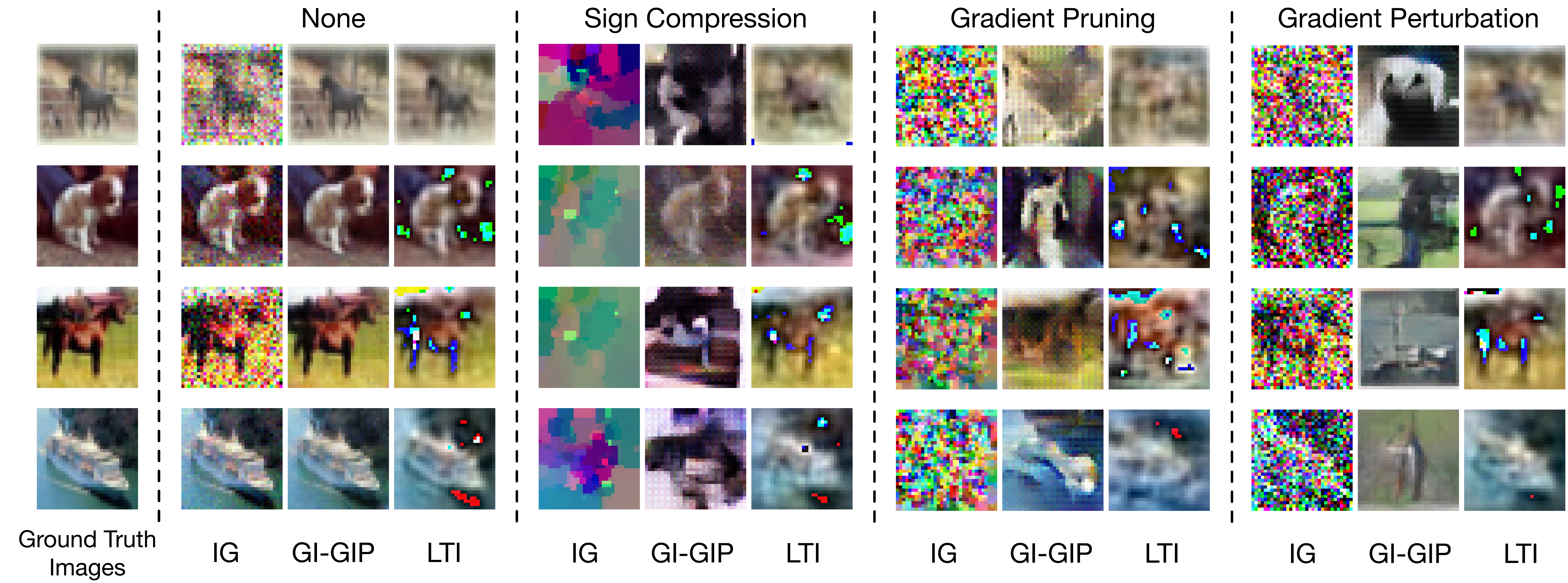}
    \caption{Comparison of LTI with IG and GI-GIP for reconstructing 4 random images in CIFAR10 when the FL model is LeNet and $B=1$. Under sign compression, only LTI can partially reconstruct the images to recover the object of interest whereas both IG and GI-GIP fail to do so on most samples.}
    \label{fig:img_example}
\end{figure*}

\subsubsection{Main Results}

\paragraph{Quantitative evaluation.} 
\autoref{tab:res_cv} gives quantitative comparisons in the metric of MSE for IG, GI-GIP, and LTI against various defense mechanisms on CIFAR10; Tables of PSNR, LPIPS and SSIM are in the appendix due to space limit. When no defense mechanism is applied, GI-GIP achieves the best performance. It is not surprising because GI-GIP, explicitly encodes image-prior in an image generator, which is more tailored than LTI to image data.
However, when the gradient is augmented with a defense mechanism that is underexplored, both IG and GI-GIP have considerably worse performance with MSE close to or above $0.1$. By comparison, LTI outperforms both baselines significantly and consistently across all three defense mechanisms.
For example, under gradient perturbation with $\sigma=0.1$, which prior work believed is sufficient for preventing gradient inversion attacks~\citep{zhu2019deep, jeon2021gradient}, MSE can be as low as $0.012$ for LTI. Our result, therefore, provides considerable additional insight for the level of empirical privacy achieved by DP-SGD~\citep{abadi2016deep}, and suggests that the theoretical privacy leakage as predicted by DP $\epsilon$ may be tighter than previously thought.
These results validate that LTI has strong adaptation performance in various settings and can be a great baseline to show the vulnerability in those underexplored settings.

\paragraph{Qualitative evaluation.} \autoref{fig:img_example} shows 4 random CIFAR10 test samples and their reconstructions under different defense mechanisms when the FL model is LeNet and $B=1$. Without any defense in place, all three methods recover a considerable amount of semantic information about the object of interest, with both GI-GIP and LTI faithfully reconstructing the training sample. Under the sign compression defense, IG completely fails to reconstruct all 4 samples, while GI-GIP only successfully reconstructs the second image. In contrast, LTI is able to recover the semantic information in all 4 samples. Results for gradient pruning and gradient perturbation yield similar conclusions. More examples are given in the appendix.

\subsubsection{Ablation Studies for Auxiliary Dataset}
\label{sec:ablation_cv}

Since LTI learns to invert gradients using the auxiliary dataset, its performance depends on the quantity and quality of data available to the adversary. We perform ablation studies to better understand this dependence by changing the auxiliary dataset size and its distribution. All ablation studies are conducted in the setting where the FL model is LeNet and $B=1$.

\begin{figure*}[t!]
    \centering
    \includegraphics[width=0.8\linewidth]{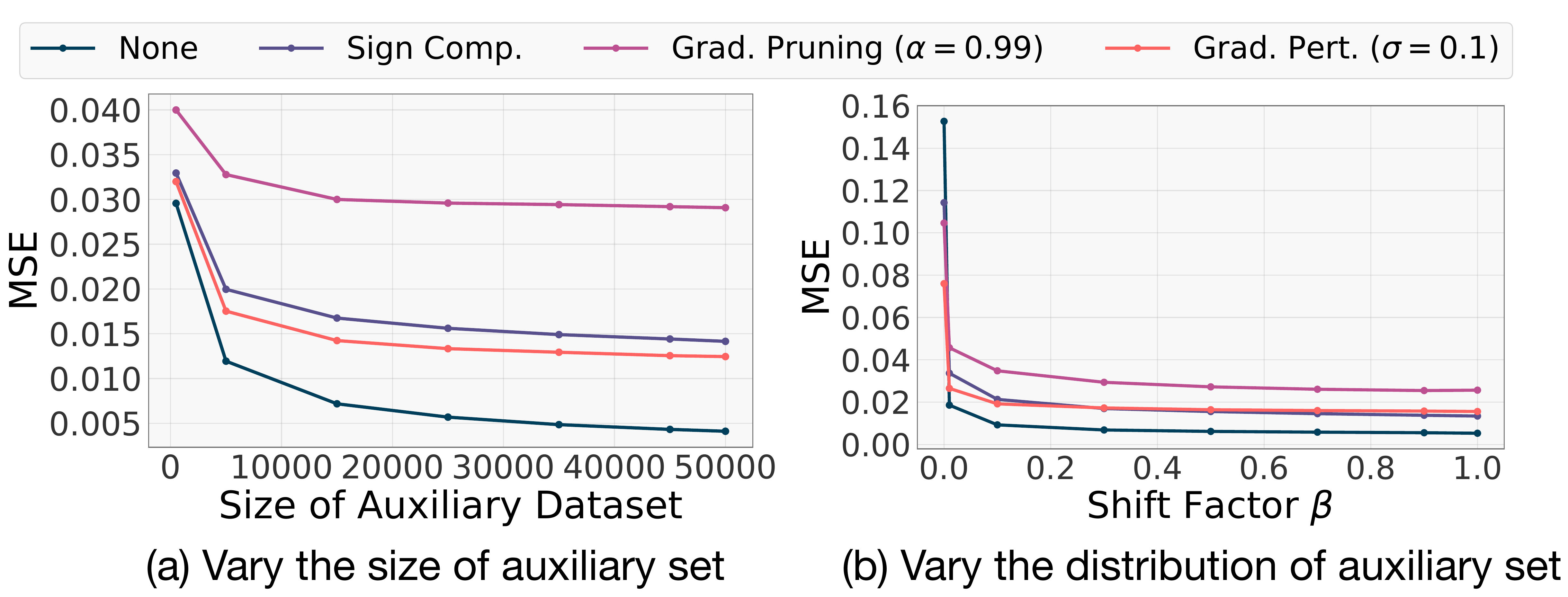}
    \caption{Ablation studies on size and distribution of the auxiliary dataset $\calD_{\rm aux}$. Under both severe data size limitation (left) and data distribution shift ($\beta=0.01$; right), LTI is able to outperform both baselines in \autoref{tab:res_cv} when a defense is applied.}
    \label{fig:ablation}
\end{figure*}

\paragraph{Varying the auxiliary dataset size.}
We randomly subsample the CIFAR10 training set to construct auxiliary datasets of size $\{500, 5000, 15000, 25000, 35000, 45000, 50000\}$ and evaluate the performance of LTI under various defenses. \autoref{fig:ablation}(a) plots reconstruction MSE as a function of the auxiliary dataset size, which is monotonically decreasing as expected. Moreover, with just $5,000$ samples for training the inversion model (second point in each curve), the performance is nearly as good as when training using the full CIFAR10 training set.
Notably, even with the auxiliary dataset size as small as $500$, the reconstruction MSE is \emph{still lower than that of IG and GI-GIP} in \autoref{tab:res_cv}.
Corresponding figures for PSNR, LPIPS, and SSIM in the appendix show similar findings.

\paragraph{Varying the auxiliary data distribution.}
Although access to a large set of in-distribution data may be unavailable in practice, the adversary may still collect out-of-distribution samples for the auxiliary dataset. This is beneficial for the adversary since a model learning on out-of-distribution samples may transfer its knowledge to in-distribution data as well. To simulate this scenario, we divide CIFAR10 into two halves with disjoint classes and construct the auxiliary dataset by combining a $\beta$ fraction of samples from the first half and a $1-\beta$ fraction of samples from the second half for $\beta \in \{0, 0.01, 0.1, 0.3, 0.5, 0.7, 0.9, 1\}$.
The target model $f_{\bw}$ is trained only on samples from the first half, and hence the auxiliary set has the exact same distribution as the target model's data when $\beta = 1$ and only has out-of-distribution data when $\beta=0$.

\autoref{fig:ablation}(b) shows reconstruction MSE as a function of $\beta$. We make the following observations: 
\begin{enumerate}[leftmargin=*,nosep]
    \item Even if the auxiliary dataset only contains $250$ in-distribution samples ($\beta = 0.01$; second point in each curve), MSE of the inversion model is \emph{still lower than that of the best baseline} in \autoref{tab:res_cv}. For example, with the sign compression defense, LTI attains an MSE of $\leq 0.02$, which is much lower than the MSE of $0.116$ for IG and $0.091$ for GI-GIP. 
    \item When the auxiliary dataset contains only out-of-distribution data ($\beta=0$), the inversion model has a very high reconstruction MSE. In the next paragraph, we will propose a data augmentation method to improve the out-of-distribution generalization.
\end{enumerate}

\begin{table}
\centering
\caption{MSE of LTI when the auxiliary dataset is out-of-distribution. LIT-OOD outperforms GI-GIP for all defense mechanism settings.}
	\label{tab:ood}
\resizebox{\linewidth}{!}{
	\begin{tabular}{c|cccc}
	\toprule
		& None & Sign Comp. & Grad. Prune. & Gauss. Pert.\\
	\midrule
		LTI-OOD & 0.015 & 0.036 & 0.045 & 0.029\\
		GI-GIP & 0.001 & 0.091 & 0.043 & 0.124\\
	\bottomrule
	\end{tabular}
	}
	
\end{table}

\paragraph{Out-of-distribution (OOD) auxiliary data.} We further consider the auxiliary dataset that only has out-of-distribution data. Suppose the auxiliary data are images of the second half classes in CIFAR10 and the target model $f_{\bw}$ is trained only on images from the first half (i.e. the setting of $\beta=0$ when studying the data distribution). Instead of performing LTI with only the out-of-distribution data, we further augment the auxiliary dataset with the following steps:
\begin{enumerate}[leftmargin=*,nosep]
	\item Convert OOD data into the frequency domain by the discrete cosine transform (DCT).
	\item Compute the mean and variance of OOD data in the DCT space.
	\item Sample new data from a Gaussian with the mean and variance computed in step 2.
	\item Convert new data back to the original image space.
\end{enumerate}
Then we can train LTI with the OOD data and the augmented data from the steps above and name this method as LTI-OOD.
Table \ref{tab:ood} presents its MSE.
By comparing it with baselines in Table \ref{tab:res_cv}, LTI-OOD is better or not worse than the baselines when the defense mechanisms are applied.
Although LTI-OOD is worse than GI-GIP when no defense mechanism is applied, this is fair because GI-GIP utilizes the in-distribution data and this is a stronger data assumption than LTI-OOD.

To better understand this data augmentation, we also test the data augmentation where we estimate a Gaussian in the original image space and the MSE will increase from 0.015 to 0.045 when no defense is applied.
We hypothesize this is because by fitting a Gaussian in the DCT domain, the frequency property as an image is kept so that the distribution is closer to the target image distribution.

\subsection{Evaluation on Language Task}
\label{sec:exp_language}

\paragraph{Federate learning tasks.} For the evaluation on language data, we consider two common language tasks: text classifier training and causal language model training\footnote{We follow the task setup and code in \hyperlink{https://github.com/JonasGeiping/breaching} {https://github.com/JonasGeiping/breaching}}. 

In the task of text classification, the classifier $f_{\bw}$ is the BERT model~\citep{devlin-etal-2019-bert} with \emph{frozen token embedding layer}. Fixing the token embedding layer is a common technique for language model fine-tuning~\citep{sun2019fine}, which also has privacy benefits since direct privacy leakage from the gradient magnitude of the token embedding layer can be prevented~\citep{fowl2022decepticons, gupta2022recovering}.
As a result, the trainable model contains about $86M$ parameters.
The BERT classier is trained on CoLA~\citep{warstadt2018neural} dataset using the cross-entropy loss.

In the task of causal language model, the language model $f_{\bw}$ is a three-layer transformer~\citep{vaswani2017attention} with \emph{frozen token embedding layer}.  The trainable model contains about $1.1M$ parameters.
We train the language model on WikiText~\citep{merity2016pointer}, where each training sample is limited to $L=16$ tokens and the language model is trained to predict the next token $\bx_l$ given $\bx_{:l-1}$ for $l=1,\ldots,L$ using the cross-entropy loss.

We set the \emph{original test split of CoLA / WikiText dataset} as the dataset for the FL training, i.e. the dataset that the attacks will be test on.

\paragraph{Defense mechanisms set-up.} The adversary will receive the gradient applied with no defense, sign compression, gradient pruning ($\alpha=0.99$) and gaussian perturbation ($\sigma=0.001$ for text classificatier training task and $\sigma=0.01$ for causal language model training task) when  $B=1$.

\paragraph{Baseline.}
We compare LTI with TAG~\citep{deng2021tag}---the state-of-the-art language model gradient inversion attack without utilizing the token embedding layer gradient\footnote{We do not compare against a more recent attack by \citet{gupta2022recovering} since it crucially depends on access to the token embedding layer gradient.}. 
The objective function for TAG is a slight modification of \autoref{eq:opt_objective} that uses both the $\ell_2$ and $\ell_1$ distance between the observed gradient and the gradient of dummy data. We also modify TAG slightly to adapt it to different defenses; see appendix for details.

\paragraph{Set-up of LTI.} We follow the setup below for training the gradient inversion model $g_\theta$.
\begin{itemize}[leftmargin=*,nosep]
    \item \textit{Auxiliary dataset.} We use 8551 samples from the train split of CoLA or $\sim 1.8\times 10^5$ samples from the train split of Wikitext as the auxiliary dataset. 
    \item \textit{Inversion model architecture.}
    For both FL tasks, we train a two-layer MLP with ReLU activation and first hidden-layer size $600$ and second hidden-layer size $1,000$. The inversion model outputs $L$ probability vectors each with size equal to the vocabulary size ($\sim 50,000$), and we train it using the cross-entropy loss to predict the $L$ tokens given the target model gradient.
    We use feature hashing (see \autoref{sec: method}) to reduce the target model gradient to $1\%$ or $10\%$ of its original dimensions as input to the inversion model when $f_{\bw}$ is BERT or three-layer transformer.
    \item \textit{Training details.} We use Adam~\citep{kingma2014adam} to train the inversion model over $100$ epochs with batch size $64$. Learning rates are selected separately for each defense from $\{10^{-3}, 10^{-4}, 10^{-5}\}$.
    \item \textit{Computation cost.} Our experiments are conducted using NVIDIA GeForce RTX 3090 GPUs and each training run takes about 3 hours.
\end{itemize}

\paragraph{Evaluation methodology.} We evaluate LTI and the TAG baseline on $1,000$ samples from each task. To measure the quality of inverted text from attacks, we use four metrics: 1. \emph{Accuracy$(\%)$} measures the average token-wise zero-one accuracy. 2. \emph{Rouge-1$(\%)$}, \emph{Rouge-2$(\%)$} and \emph{Rouge-L$(\%)$} measure the overlap of unigram, bigram, and length of longest common subsequence between the ground truth and the reconstructed text. 

We also check the reconstructed texts from both TAG and LTI to see how the semantic meaning of the text is recovered and analyze the type of reconstruction error. This part is put in the appendix.

\begin{table*}[t!]
\centering
\caption{Results for gradient inversion attack on two language tasks. The overall trend is remarkably consistent: in all 4 metrics, LTI significantly outperforms TAG across different settings (7 out of 8). This shows that our method is easily adapted and is able to achieve great attack performance.}
    \label{tab:res_nlp}
	\subfloat[{\normalsize Text classifier training on CoLA dataset.}]{
    \centering
    \begin{tabular}{c|cccc|cccc}
    \toprule
        \textbf{Defense}& \multicolumn{4}{c|}{\textbf{None}} & \multicolumn{4}{c}{\textbf{Sign Compression}} \\
        \midrule
         \textbf{Method}  & Acc.  & Rouge-1 & Rouge-2 & Rouge-L & Acc.  & Rouge-1 & Rouge-2 & Rouge-L  \\
        \midrule
          TAG &  $      8.38$ & $     51.23$ & $      6.88$ & $     29.35$&$      1.62$ & $      8.81$ & $      0.00$ & $      8.09$  \\
          LTI (Ours) &  $     61.87$&$     65.23$&$     44.46$&$     63.34$& $     63.89$&$     69.92$&$     49.79$&$     67.86$  \\
          \midrule
          LTI-OOD (Ours)&	$     52.03$&$     45.86$&$     29.46$&$     45.79$&$     50.77$&$     49.07$&$     30.86$&$     48.80$ \\
     \midrule
     \midrule
         \textbf{Defense}&  \multicolumn{4}{c|}{\textbf{Gradient Pruning} ($\alpha=0.99$)} & \multicolumn{4}{c}{\textbf{Gaussian Perturbation} ($\sigma=0.001$)} \\
        \midrule
         \textbf{Method}  & Acc.  & Rouge-1 & Rouge-2 & Rouge-L & Acc.  & Rouge-1 & Rouge-2 & Rouge-L  \\
        \midrule
         TAG &  $      5.69$ & $     43.30$ & $      6.90$ & $     26.96$&$      5.12$ & $     33.85$ & $      2.94$ & $     22.01$ \\
         LTI (Ours) & $     58.93$&$     60.12$&$     37.96$&$     58.17$&$     53.96$&$     53.09$&$     32.41$&$     52.35$\\
         \midrule 
         LTI-OOD (Ours) & $     38.68$&$     35.66$&$     23.11$&$     35.46$& $     37.96$&$     33.75$&$     21.85$&$     33.55$\\
     \bottomrule
    \end{tabular}%
	}
	\\
	\subfloat[{\normalsize Causal language model training on WikiText dataset.}]{
    \centering
    \begin{tabular}{c|cccc|cccc}
    \toprule
        \textbf{Defense}& \multicolumn{4}{c|}{\textbf{None}} & \multicolumn{4}{c}{\textbf{Sign Compression}} \\
        \midrule
        \textbf{Method}  & Acc.  & Rouge-1 & Rouge-2 & Rouge-L & Acc.  & Rouge-1 & Rouge-2 & Rouge-L  \\
        \midrule
         TAG &   $     74.13$ & $     71.92$ & $     50.64$ & $     68.46$& $    100.00$ & $    100.00$ & $    100.00$ & $    100.00$   \\
         LTI (Ours) &  $     89.61$&$     86.91$&$     80.68$&$     86.90$& $71.15$&$     64.35$&$     45.40$&$     64.29$  \\
         \midrule 
         LTI-OOD (Ours) &   $     91.14$&$     89.43$&$     85.11$&$     89.41$& $     88.06$&$     84.66$&$     76.46$&$     84.64$  \\
     \midrule
     \midrule
        \textbf{Defense}&  \multicolumn{4}{c|}{\textbf{Gradient Pruning} ($\alpha=0.99$)} & \multicolumn{4}{c}{\textbf{Gaussian Perturbation} ($\sigma=0.01$)} \\
        \midrule
        \textbf{Method}  & Acc.  & Rouge-1 & Rouge-2 & Rouge-L & Acc.  & Rouge-1 & Rouge-2 & Rouge-L  \\
        \midrule
        TAG & $     34.34$ & $     48.50$ & $     10.21$ & $     35.60$ & $     64.34$ & $     66.19$ & $     37.86$ & $     59.55$  \\
        LTI (Ours) & $     70.80$&$     64.24$&$     45.79$&$     64.15$& $     82.49$&$     78.75$&$     67.06$&$     78.71$\\
        \midrule 
        LTI-OOD (Ours) & $     86.19$&$     82.56$&$     73.04$&$     82.50$& $     90.25$&$     87.39$&$     81.94$&$     87.34$\\
     \bottomrule
    \end{tabular}%
	}
    
\end{table*}

\paragraph{Results.} 
\autoref{tab:res_nlp} shows the quantitative comparison between LTI and TAG against various defenses. The overall trend is remarkably consistent: in all 4 metrics, LTI significantly outperforms TAG across different settings (7 out of 8). This shows that our method is easily adapted to the discrete language data and different defenses and is able to achieve great attack performance. 

One observation is that the accuracy of inverted texts when the FL task is the causal language model training is overall much higher than the accuracy when the FL task is the text classifier training. We hypothesize this is because in the task of causal language model, the label in the cross entropy loss is the input sequence itself. On the other hand, The literature \citep{yin2021see, zhao2020idlg} shows how easy it is to reconstruct the labels. 

Another observation is that TAG has a relative low performance at most settings, it achieves the perfect accuracy at the setting of the sign compression when the FL task is the causal language model training. At the first impression, this perfect accuracy is very suspicious. By our carefully check, the explanation is that: if we treat the objective function when the gradient is applied sign compression as a special objective function when the adversary receives the full gradient, the result simply suggests that this special objective function is coincidently better than the one designed for the full gradient. Nevertheless, this phenomenon is not generalized to TAG for the other FL task. This demonstrates that the optimization-based method is very sensitive to the design of the object function.

\paragraph{Out-of-distribution (OOD) auxiliary data.} Instead of assuming the adversary has in-distribution auxiliary texts, we relax this to only assuming the knowledge of the word frequency. Then, we can independently sample the word token for each position in the sentence and get a set of pseudo data. The distribution of pseudo data is out-of-distribution, because the pseudo data loses the inner dependency between different positions of a sentence. We train LTI with the pseudo data and name it as LTI-OOD. 

The results of LTI-OOD are presented in Table~\ref{tab:res_nlp}. LTI outperforms TAG on both CoLA and WikiText dataset at most metrics for all settings of gradient defenses. Moreover, we can observe that LTI-OOD is even better than LTI on WikiText dataset. We hope this promising OOD results can motivate the exploration of OOD generalization of LTI in the future work.



\section{Conclusion and Future Work}
\label{sec:conclusion}

We demonstrated the effectiveness of LTI---a simple learning-based gradient inversion attack---under realistic federated learning settings. For both vision and language tasks, LTI can match or exceed the performance of state-of-the-art optimization-based methods when no defense is applied, and significantly outperform all prior works under defenses based on gradient perturbation and gradient compression. Given its simplicity and versatility, we advocate the use of LTI as both a strong baseline for future research and a diagnostic tool for evaluating privacy leakage in FL.


\textbf{Future work.} This paper serves as preliminary work towards understanding the effectiveness of learning-based gradient inversion attacks, and our method can be further improved in several directions. 
\textbf{1.} For large models, our current approach is to hash the gradients into a lower-dimensional space to reduce memory cost. It may be possible to leverage model architectures to design more effective dimensionality reduction techniques to further scale up the method. 
\textbf{2.} Currently we only focus on the setting with batch size 4 for vision tasks and batch size 1 for language tasks. In practice, the batch size could be larger. For LTI, the complexity of MLP would increase when the batch size increases, which makes learning harder. More advanced model architectures and loss designs may help with the large batch case.
\textbf{3.} LTI in its current form does not leverage additional data priors such as image smoothness or text fluency. We can readily incorporate these priors by modifying the inversion model's loss function with total variation (for image data) or perplexity on a trained language model (for text data), which may further improve the performance of LTI. 

\begin{acknowledgements}
	RW, XC and KQW are supported by grants from the National Science Foundation NSF (IIS-2107161, III-1526012, IIS-1149882, and IIS-1724282), and the Cornell Center for Materials Research with funding from the NSF MRSEC program (DMR-1719875), and SAP America. RW is also supported by LinkedIn PhD Award.
\end{acknowledgements}

\bibliography{citations}

\begin{thebibliography}{32}
\providecommand{\natexlab}[1]{#1}
\providecommand{\url}[1]{\texttt{#1}}
\expandafter\ifx\csname urlstyle\endcsname\relax
  \providecommand{\doi}[1]{doi: #1}\else
  \providecommand{\doi}{doi: \begingroup \urlstyle{rm}\Url}\fi

\bibitem[Abadi et~al.(2016)Abadi, Chu, Goodfellow, McMahan, Mironov, Talwar,
  and Zhang]{abadi2016deep}
Martin Abadi, Andy Chu, Ian Goodfellow, H~Brendan McMahan, Ilya Mironov, Kunal
  Talwar, and Li~Zhang.
\newblock Deep learning with differential privacy.
\newblock In \emph{Proceedings of the 2016 ACM SIGSAC conference on computer
  and communications security}, pages 308--318, 2016.

\bibitem[Aji and Heafield(2017)]{aji2017sparse}
Alham~Fikri Aji and Kenneth Heafield.
\newblock Sparse communication for distributed gradient descent.
\newblock \emph{arXiv preprint arXiv:1704.05021}, 2017.

\bibitem[Bernstein et~al.(2018)Bernstein, Wang, Azizzadenesheli, and
  Anandkumar]{bernstein2018signsgd}
Jeremy Bernstein, Yu-Xiang Wang, Kamyar Azizzadenesheli, and Animashree
  Anandkumar.
\newblock signsgd: Compressed optimisation for non-convex problems.
\newblock In \emph{International Conference on Machine Learning}, pages
  560--569. PMLR, 2018.

\bibitem[Bishop et~al.(1995)]{bishop1995neural}
Christopher~M Bishop et~al.
\newblock \emph{Neural networks for pattern recognition}.
\newblock Oxford university press, 1995.

\bibitem[Boenisch et~al.(2021)Boenisch, Dziedzic, Schuster, Shamsabadi,
  Shumailov, and Papernot]{boenisch2021curious}
Franziska Boenisch, Adam Dziedzic, Roei Schuster, Ali~Shahin Shamsabadi, Ilia
  Shumailov, and Nicolas Papernot.
\newblock When the curious abandon honesty: Federated learning is not private.
\newblock \emph{arXiv preprint arXiv:2112.02918}, 2021.

\bibitem[Bonawitz et~al.(2016)Bonawitz, Ivanov, Kreuter, Marcedone, McMahan,
  Patel, Ramage, Segal, and Seth]{bonawitz2016practical}
Keith Bonawitz, Vladimir Ivanov, Ben Kreuter, Antonio Marcedone, H~Brendan
  McMahan, Sarvar Patel, Daniel Ramage, Aaron Segal, and Karn Seth.
\newblock Practical secure aggregation for federated learning on user-held
  data.
\newblock \emph{arXiv preprint arXiv:1611.04482}, 2016.

\bibitem[Deng et~al.(2021)Deng, Wang, Li, Shang, Liu, Rajasekaran, and
  Ding]{deng2021tag}
Jieren Deng, Yijue Wang, Ji~Li, Chao Shang, Hang Liu, Sanguthevar Rajasekaran,
  and Caiwen Ding.
\newblock Tag: Gradient attack on transformer-based language models.
\newblock \emph{arXiv preprint arXiv:2103.06819}, 2021.

\bibitem[Devlin et~al.(2019)Devlin, Chang, Lee, and
  Toutanova]{devlin-etal-2019-bert}
Jacob Devlin, Ming-Wei Chang, Kenton Lee, and Kristina Toutanova.
\newblock {BERT}: Pre-training of deep bidirectional transformers for language
  understanding.
\newblock In \emph{Conference of the North {A}merican Chapter of the
  Association for Computational Linguistics}, pages 4171--4186, Minneapolis,
  Minnesota, 2019. Association for Computational Linguistics.

\bibitem[Dwork et~al.(2006)Dwork, McSherry, Nissim, and
  Smith]{dwork2006calibrating}
Cynthia Dwork, Frank McSherry, Kobbi Nissim, and Adam Smith.
\newblock Calibrating noise to sensitivity in private data analysis.
\newblock In \emph{Theory of cryptography conference}, pages 265--284.
  Springer, 2006.

\bibitem[Dwork et~al.(2014)Dwork, Roth, et~al.]{dwork2014algorithmic}
Cynthia Dwork, Aaron Roth, et~al.
\newblock The algorithmic foundations of differential privacy.
\newblock \emph{Found. Trends Theor. Comput. Sci.}, 9\penalty0 (3-4):\penalty0
  211--407, 2014.

\bibitem[Fowl et~al.(2021)Fowl, Geiping, Czaja, Goldblum, and
  Goldstein]{fowl2021robbing}
Liam Fowl, Jonas Geiping, Wojtek Czaja, Micah Goldblum, and Tom Goldstein.
\newblock Robbing the fed: Directly obtaining private data in federated
  learning with modified models.
\newblock \emph{arXiv preprint arXiv:2110.13057}, 2021.

\bibitem[Fowl et~al.(2022)Fowl, Geiping, Reich, Wen, Czaja, Goldblum, and
  Goldstein]{fowl2022decepticons}
Liam Fowl, Jonas Geiping, Steven Reich, Yuxin Wen, Wojtek Czaja, Micah
  Goldblum, and Tom Goldstein.
\newblock Decepticons: Corrupted transformers breach privacy in federated
  learning for language models.
\newblock \emph{arXiv preprint arXiv:2201.12675}, 2022.

\bibitem[Geiping et~al.(2020)Geiping, Bauermeister, Dr{\"o}ge, and
  Moeller]{geiping2020inverting}
Jonas Geiping, Hartmut Bauermeister, Hannah Dr{\"o}ge, and Michael Moeller.
\newblock Inverting gradients-how easy is it to break privacy in federated
  learning?
\newblock \emph{Advances in Neural Information Processing Systems},
  33:\penalty0 16937--16947, 2020.

\bibitem[Gupta et~al.(2022)Gupta, Huang, Zhong, Gao, Li, and
  Chen]{gupta2022recovering}
Samyak Gupta, Yangsibo Huang, Zexuan Zhong, Tianyu Gao, Kai Li, and Danqi Chen.
\newblock Recovering private text in federated learning of language models.
\newblock \emph{arXiv preprint arXiv:2205.08514}, 2022.

\bibitem[He et~al.(2016)He, Zhang, Ren, and Sun]{he2016deep}
Kaiming He, Xiangyu Zhang, Shaoqing Ren, and Jian Sun.
\newblock Deep residual learning for image recognition.
\newblock In \emph{Proceedings of the IEEE conference on computer vision and
  pattern recognition}, pages 770--778, 2016.

\bibitem[Jeon et~al.(2021)Jeon, Lee, Oh, Ok, et~al.]{jeon2021gradient}
Jinwoo Jeon, Kangwook Lee, Sewoong Oh, Jungseul Ok, et~al.
\newblock Gradient inversion with generative image prior.
\newblock \emph{Advances in Neural Information Processing Systems},
  34:\penalty0 29898--29908, 2021.

\bibitem[Kasiviswanathan et~al.(2011)Kasiviswanathan, Lee, Nissim,
  Raskhodnikova, and Smith]{kasiviswanathan2011can}
Shiva~Prasad Kasiviswanathan, Homin~K Lee, Kobbi Nissim, Sofya Raskhodnikova,
  and Adam Smith.
\newblock What can we learn privately?
\newblock \emph{SIAM Journal on Computing}, 40\penalty0 (3):\penalty0 793--826,
  2011.

\bibitem[Kingma and Ba(2014)]{kingma2014adam}
Diederik~P Kingma and Jimmy Ba.
\newblock Adam: A method for stochastic optimization.
\newblock \emph{arXiv preprint arXiv:1412.6980}, 2014.

\bibitem[Krizhevsky et~al.(2009)Krizhevsky, Hinton,
  et~al.]{krizhevsky2009learning}
Alex Krizhevsky, Geoffrey Hinton, et~al.
\newblock Learning multiple layers of features from tiny images.
\newblock 2009.

\bibitem[LeCun et~al.(1998)LeCun, Bottou, Bengio, and
  Haffner]{lecun1998gradient}
Yann LeCun, L{\'e}on Bottou, Yoshua Bengio, and Patrick Haffner.
\newblock Gradient-based learning applied to document recognition.
\newblock \emph{Proceedings of the IEEE}, 86\penalty0 (11):\penalty0
  2278--2324, 1998.

\bibitem[McMahan et~al.(2017)McMahan, Moore, Ramage, Hampson, and
  y~Arcas]{mcmahan2017communication}
Brendan McMahan, Eider Moore, Daniel Ramage, Seth Hampson, and Blaise~Aguera
  y~Arcas.
\newblock Communication-efficient learning of deep networks from decentralized
  data.
\newblock In \emph{Artificial intelligence and statistics}, pages 1273--1282.
  PMLR, 2017.

\bibitem[Merity et~al.(2016)Merity, Xiong, Bradbury, and
  Socher]{merity2016pointer}
Stephen Merity, Caiming Xiong, James Bradbury, and Richard Socher.
\newblock Pointer sentinel mixture models, 2016.

\bibitem[Shokri et~al.(2017)Shokri, Stronati, Song, and
  Shmatikov]{shokri2017membership}
Reza Shokri, Marco Stronati, Congzheng Song, and Vitaly Shmatikov.
\newblock Membership inference attacks against machine learning models.
\newblock In \emph{2017 IEEE symposium on security and privacy (SP)}, pages
  3--18. IEEE, 2017.

\bibitem[Simonyan and Zisserman(2014)]{simonyan2014very}
Karen Simonyan and Andrew Zisserman.
\newblock Very deep convolutional networks for large-scale image recognition.
\newblock \emph{arXiv preprint arXiv:1409.1556}, 2014.

\bibitem[Sun et~al.(2019)Sun, Qiu, Xu, and Huang]{sun2019fine}
Chi Sun, Xipeng Qiu, Yige Xu, and Xuanjing Huang.
\newblock How to fine-tune bert for text classification?
\newblock In \emph{China national conference on Chinese computational
  linguistics}, pages 194--206. Springer, 2019.

\bibitem[Vaswani et~al.(2017)Vaswani, Shazeer, Parmar, Uszkoreit, Jones, Gomez,
  Kaiser, and Polosukhin]{vaswani2017attention}
Ashish Vaswani, Noam Shazeer, Niki Parmar, Jakob Uszkoreit, Llion Jones,
  Aidan~N Gomez, {\L}ukasz Kaiser, and Illia Polosukhin.
\newblock Attention is all you need.
\newblock \emph{Advances in neural information processing systems}, 30, 2017.

\bibitem[Warstadt et~al.(2018)Warstadt, Singh, and Bowman]{warstadt2018neural}
Alex Warstadt, Amanpreet Singh, and Samuel~R Bowman.
\newblock Neural network acceptability judgments.
\newblock \emph{arXiv preprint arXiv:1805.12471}, 2018.

\bibitem[Weinberger et~al.(2009)Weinberger, Dasgupta, Langford, Smola, and
  Attenberg]{weinberger2009feature}
Kilian Weinberger, Anirban Dasgupta, John Langford, Alex Smola, and Josh
  Attenberg.
\newblock Feature hashing for large scale multitask learning.
\newblock In \emph{Proceedings of the 26th annual international conference on
  machine learning}, pages 1113--1120, 2009.

\bibitem[Yin et~al.(2021)Yin, Mallya, Vahdat, Alvarez, Kautz, and
  Molchanov]{yin2021see}
Hongxu Yin, Arun Mallya, Arash Vahdat, Jose~M Alvarez, Jan Kautz, and Pavlo
  Molchanov.
\newblock See through gradients: Image batch recovery via gradinversion.
\newblock In \emph{Proceedings of the IEEE/CVF Conference on Computer Vision
  and Pattern Recognition}, pages 16337--16346, 2021.

\bibitem[Zhang et~al.(2019)Zhang, Hare, and Prugel-Bennett]{zhang2019deep}
Yan Zhang, Jonathon Hare, and Adam Prugel-Bennett.
\newblock Deep set prediction networks.
\newblock \emph{Advances in Neural Information Processing Systems}, 32, 2019.

\bibitem[Zhao et~al.(2020)Zhao, Mopuri, and Bilen]{zhao2020idlg}
Bo~Zhao, Konda~Reddy Mopuri, and Hakan Bilen.
\newblock idlg: Improved deep leakage from gradients.
\newblock \emph{arXiv preprint arXiv:2001.02610}, 2020.

\bibitem[Zhu et~al.(2019)Zhu, Liu, and Han]{zhu2019deep}
Ligeng Zhu, Zhijian Liu, and Song Han.
\newblock Deep leakage from gradients.
\newblock \emph{Advances in Neural Information Processing Systems}, 32, 2019.

\end{thebibliography}

\appendix
\newpage
\onecolumn

\section{Modifications for Baseline Methods}
\paragraph{Vision baselines.} 
IG and GI-GIP use cosine distance between the received gradient and the gradient of dummy data for optimizing the dummy data. 
However, reusing this objective when defense mechanisms are applied is not reasonable.

For the \emph{sign compression} defense, this loss function does not optimize the correct objective since the dummy data's gradient is \emph{not} a vector with $\pm 1$ entries but rather a real-valued vector with the same sign.
When $B=1$, we can simiply replace cosine distance by the loss $\sum_{i=1}^m\left(\ell_{\rm sign}^i\right)^2$ where 
\begin{equation}
    \label{eq:sign_loss}
    \ell_{\rm sign}^i =\max\left\{-\nabla_{\bw_i} \ell(f_{\bw}(\tilde{\bx}), \tilde{y})\cdot \mathrm{Sign}\left( \nabla_{\bw_i} \ell(f_{\bw}(\bx), y)\right), 0\right\}.
\end{equation}
One sanity check for this loss is that when $\nabla_{\bw_i} \ell(f_{\bw}(\tilde{\bx}), \tilde{y})$ has the same sign as that of $\nabla_{\bw_i} \ell(f_{\bw}(\bx), y)$, the minimum loss value of $0$ is achieved.
When $B>1$, the above objective can't be applied anymore because the adversary only receives the average of the gradients that are compressed to sign and doesn't know the gradient sign for each single data.
Because sign operation is not reasonably differentiable, we can't compute the average of sign gradients from dummy data and reuse the cosine distance as the objective function.
However, the $tanh$ function is approximate to the sign operation and is differentiable. 
Thus, the solution is to apply $tanh$ to the gradient of each dummy data, compute the average of them, and reuse the cosine distance between this average and the received gradient.

For the \emph{gradient pruning} defense, optimizing the cosine distance between the dummy data gradient and the pruned ground truth gradient will force too many gradient values to $0$, which is the incorrect value for the ground truth gradient.
Therefore we only compute cosine distance over the non-zero dimensions of the pruned gradient. 

\paragraph{Language baselines.} For TAG, we find that the loss function also needs to be modified slightly to accommodate the \emph{sign compression} and \emph{gradient pruning} defenses:
\begin{itemize}[leftmargin=*,nosep]
    \item \textit{Sign compression.} Similar to the vision baselines, the $\ell_2$ and $\ell_1$ distance between the dummy data gradient and the ground truth gradient sign do not optimize the correct objective. When $B=1$, we can simply replace $\|\cdot\|_2^2$ and $\|\cdot\|_1$ by $\sum_{i=1}^m \left(\ell_{\rm sign}^i\right)^2 $ and $\ell_{\rm sign}^i$, respectively, where $\sum_{i=1}^m\ell_{\rm sign}^i$ is defined in \autoref{eq:sign_loss}. We make the modification similar to the vision baselines when $B>1$.
    \item \textit{Gradient pruning.} We make the same modification to TAG as in the vision baselines.
\end{itemize}

\section{Additional Quantitative Evaluation}
In the experiment of vision tasks, we evaluate the gradient inversion attacks in three metrics: MSE, PSNR, LPIPS, SSIM. In the main text, we showed the result table for MSE. \autoref{tab:res_cv_psnr}, \autoref{tab:res_cv_lpips} and \autoref{tab:res_cv_ssim} are the result tables for PSNR, LPIPS and SSIM. Similar to the trends in the MSE table, LTI is the best when the defense mechanisms are applied. 
\begin{table*}[ht!]
\centering
    \resizebox{\linewidth}{!}{
    \begin{tabular}{c|c|cccc|cccc}
    \toprule
     \multirow{2}{*}{\textbf{FL model}}  & \multirow{2}{*}{\textbf{Methods}} & \multicolumn{4}{c|}{$B=1$}& \multicolumn{4}{c}{$B=4$}\\
     \cmidrule{3-10}
         & & \textbf{None} & \textbf{Sign Comp.} & \textbf{Grad. Prun.} & \textbf{Gauss. Pert.} & \textbf{None} & \textbf{Sign Comp.} & \textbf{Grad. Prun.} & \textbf{Gauss. Pert.}\\ 
        \midrule
        \multirow{3}{*}{LeNet} & IG & 22.290 & 9.981 & 8.807 & 8.349 & 10.102 & 5.808 & 8.175 & 6.891\\ 
         & GI-GIP   & \textbf{33.374} & 13.574 & 14.356 & 9.383 &  \textbf{23.891} & 10.953 & 7.606 & 8.347\\ 
         & LTI (Ours) & 24.837 &  \textbf{18.986} &  \textbf{15.897} &  \textbf{20.249} & 19.491 &  \textbf{16.991} &  \textbf{15.643} &  \textbf{16.619}\\
         \midrule
         \multirow{3}{*}{ResNet20} & IG & 9.285 & 8.416 & 7.722 & 8.934 & 9.171 & 5.675 & 7.207 & 9.225\\ 
         & GI-GIP   & 12.609  & 10.391  & 6.286  & 6.461 & 11.064 & 6.532 & 6.562 & 6.622\\ 
         & LTI (Ours) & \textbf{18.007} & \textbf{19.435} & \textbf{16.957} & \textbf{17.367} & \textbf{12.593} & \textbf{12.290} & \textbf{12.530} & \textbf{12.613}\\
     \bottomrule
    \end{tabular}%
    }
    \caption{PSNR for baselines (IG and GI-GIP) and our method LTI on CIFAR10.} 
    \label{tab:res_cv_psnr}
\end{table*}

\begin{table*}[ht!]
\centering
    \resizebox{\linewidth}{!}{%
    \begin{tabular}{c|c|cccc|cccc}
    \toprule
      \multirow{2}{*}{\textbf{FL model}}  & \multirow{2}{*}{\textbf{Methods}} & \multicolumn{4}{c|}{$B=1$}& \multicolumn{4}{c}{$B=4$}\\
     \cmidrule{3-10}
         & & \ \textbf{None} & \textbf{Sign Comp.} & \textbf{Grad. Prun.} & \textbf{Gauss. Pert.} & \textbf{None} & \textbf{Sign Comp.} & \textbf{Grad. Prun.} & \textbf{Gauss. Pert.}\\ 
        \midrule
        \multirow{3}{*}{LeNet} & IG         & 0.263 & 0.677          & 0.675          & 0.653          & 0.615          & 0.712          & 0.690          & 0.691          \\
& GI-GIP     & \textbf{0.033}                & 0.471          & 0.474          & 0.568          & \textbf{0.212} & 0.586          & 0.695          & 0.678          \\
& LTI (Ours) & 0.221                & \textbf{0.396} & \textbf{0.472} & \textbf{0.370} & 0.391 & \textbf{0.467} & \textbf{0.489} & \textbf{0.470}\\
    \midrule
    \multirow{3}{*}{ResNet20} & IG         & 0.655 & 0.678          & 0.688          & 0.660          & 0.658          & 0.714          & 0.704          & 0.656          \\
& GI-GIP     & 0.557                & 0.650          & 0.706          & 0.701          & \textbf{0.586} & 0.671          & 0.714          & 0.712          \\
& LTI (Ours) & \textbf{0.524}                & \textbf{0.431} & \textbf{0.541} & \textbf{0.529} & 0.628 & \textbf{0.580} & \textbf{0.609} & \textbf{0.620}\\
     \bottomrule
    \end{tabular}%
    }
    \caption{LPIPS  for baselines (IG and GI-GIP) and our method LTI on CIFAR10.} 
    \label{tab:res_cv_lpips}
\end{table*}

\begin{table*}[ht!]
\centering
    \resizebox{\linewidth}{!}{%
    \begin{tabular}{c|c|cccc|cccc}
    \toprule
      \multirow{2}{*}{\textbf{FL model}}  & \multirow{2}{*}{\textbf{Methods}} & \multicolumn{4}{c|}{$B=1$}& \multicolumn{4}{c}{$B=4$}\\
     \cmidrule{3-10}
         & & \ \textbf{None} & \textbf{Sign Comp.} & \textbf{Grad. Prun.} & \textbf{Gauss. Pert.} & \textbf{None} & \textbf{Sign Comp.} & \textbf{Grad. Prun.} & \textbf{Gauss. Pert.}\\ 
        \midrule
        \multirow{3}{*}{LeNet} & IG         & 0.711         & 	0.060         & 	0.052         & 	0.149         & 	0.020         & 	0.018         & 	0.025     & 	0.058\\
& GI-GIP     & \textbf{0.970}		&	0.301	&	0.346	&	0.072	&	\textbf{0.805}	&	0.307	&	0.010	&	0.013         \\
& LTI (Ours) & 0.845	&	\textbf{0.599}	&	\textbf{0.378}	&	\textbf{0.636}	&	0.583	&	\textbf{0.432}	&	\textbf{0.330}	&	\textbf{0.425}\\
    \midrule
    \multirow{3}{*}{ResNet20} & IG         & 0.071	&	0.037	&	0.023	&	0.067		&0.046	&	0.009	&	0.018	&	0.053         \\
& GI-GIP     & 0.167		&0.049	&	0.004	&	0.008	&	0.100	&	0.034	&	0.012	&	0.012 \\
& LTI (Ours) & \textbf{0.417	}	&\textbf{0.556}	&	\textbf{0.349}	&	\textbf{0.376}	&	\textbf{0.194}	&	\textbf{0.256}	&	\textbf{0.210}	&	\textbf{0.201}\\
     \bottomrule
    \end{tabular}%
    }
    \caption{SSIM for baselines (IG and GI-GIP) and our method LTI on CIFAR10.} 
    \label{tab:res_cv_ssim}
\end{table*}

\section{Auxiliary Dataset Ablation Studies}
\begin{figure}[!t]
    \centering
    \includegraphics[width=\linewidth]{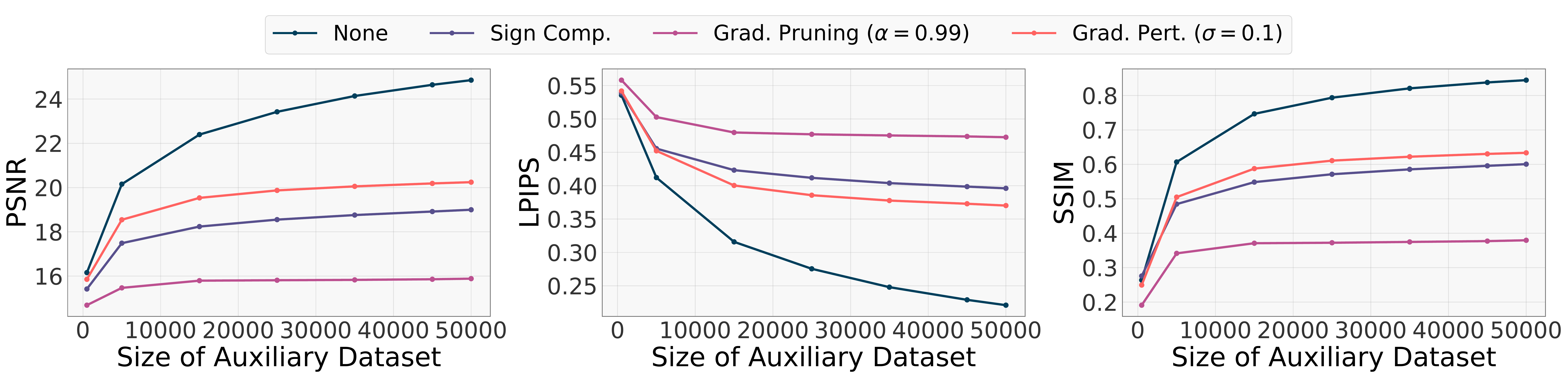}
    \caption{Plot of reconstruction PSNR / LPIPS / SSIM vs. auxiliary dataset size on CIFAR10.}
    \label{fig:ablation_appendix_size}
\end{figure}
\begin{figure}[!t]
    \centering
    \includegraphics[width=\linewidth]{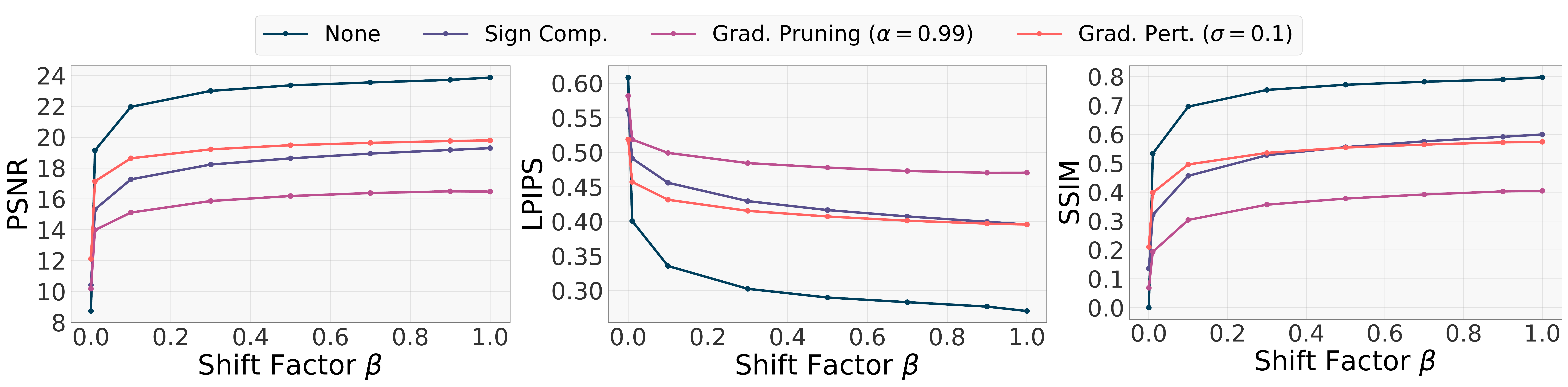}
    \caption{Plot of reconstruction PSNR / LPIPS / SSIM vs. auxiliary dataset distribution on CIFAR10.}
    \label{fig:ablation_appendix_shift}
\end{figure}

In the experiment section, we showed reconstruction MSE for LTI as a function of the auxiliary dataset size and the shift factor $\beta$. 
For completeness, we show the corresponding PSNR, LPIPS and SSIM curves in \autoref{fig:ablation_appendix_size} and \autoref{fig:ablation_appendix_shift}.
Similar to Figure 2 in the main text, when reducing the auxiliary dataset size (\emph{e.g.}, from $50,000$ to $5,000$) or reducing the proportion of in-distribution data (\emph{e.g.}, from $\beta=1$ to $\beta=0.1$), the performance of LTI does not worsen significantly.

\section{Additional Examples}

\begin{figure}[b]
    \centering
    \includegraphics[width=0.85\linewidth]{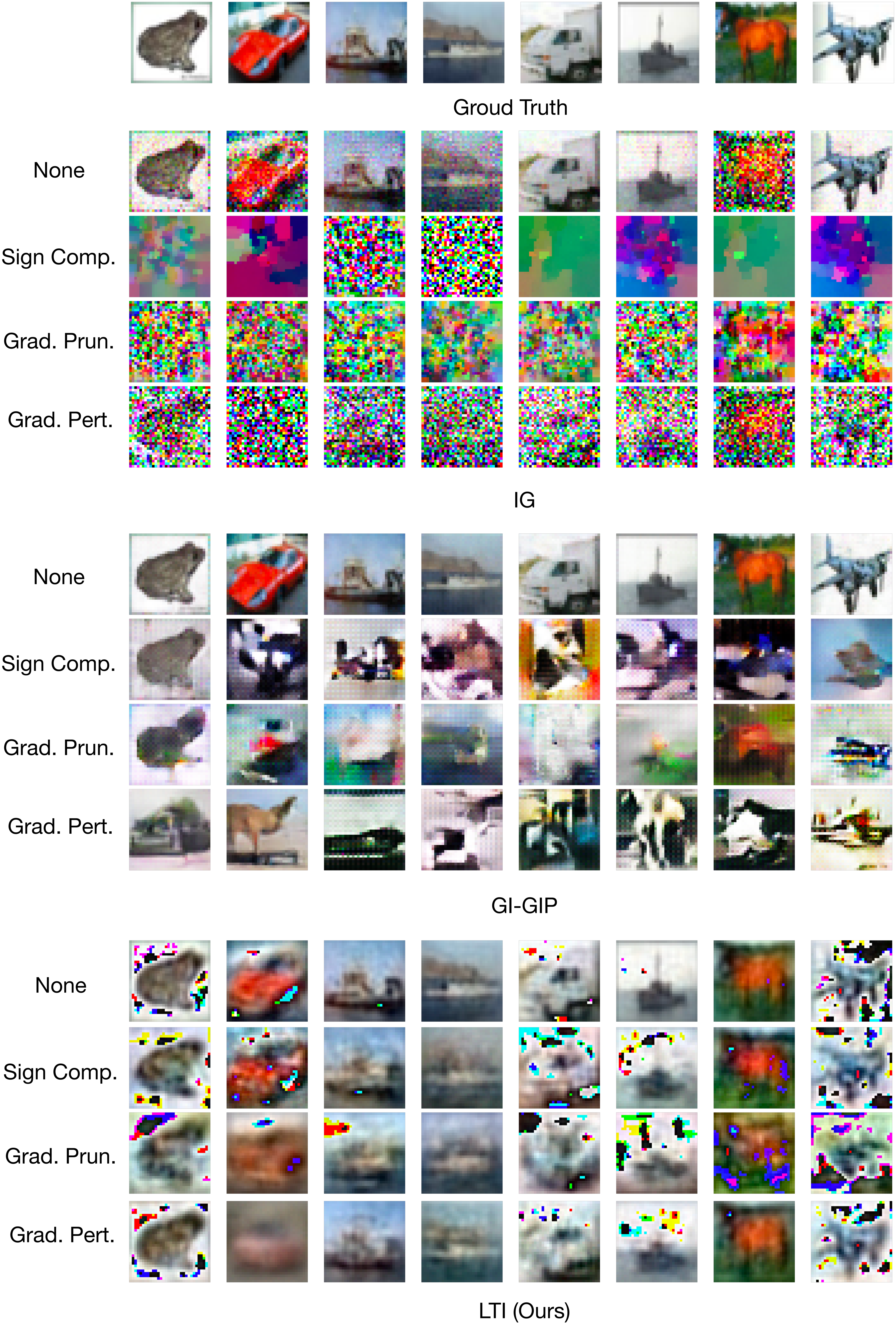}
    \caption{Additional samples from CIFAR10 and their reconstructions from the gradient of LeNet.}
    \label{fig:img_example_more}
\end{figure}
\begin{figure}[b]
    \centering
    \includegraphics[width=0.85\linewidth]{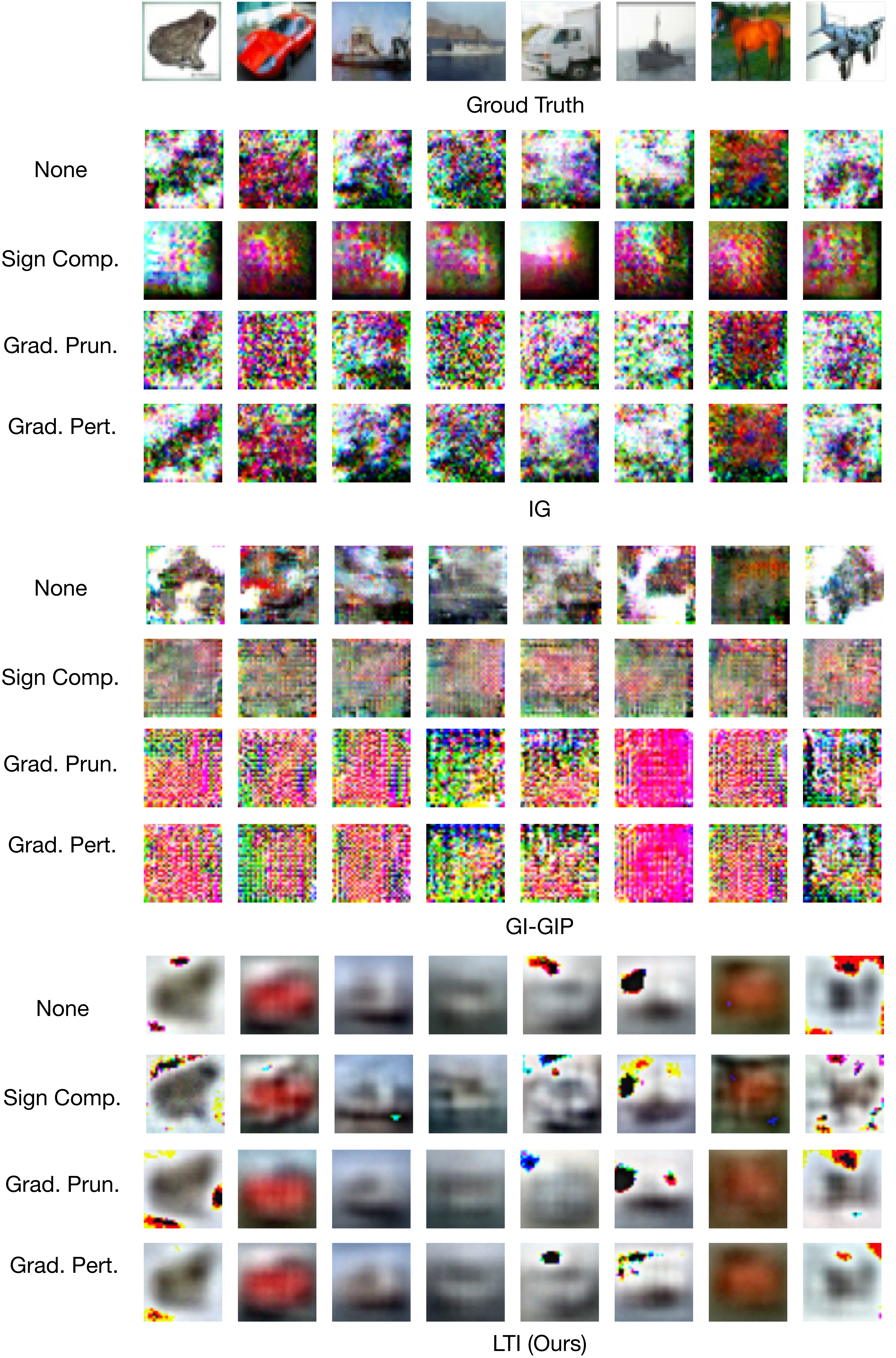}
    \caption{Additional samples from CIFAR10 and their reconstructions from the gradient of ResNet20.}
    \label{fig:img_example_resnet}
\end{figure}

\begin{figure}[h]
    \centering
    \includegraphics[width=\linewidth]{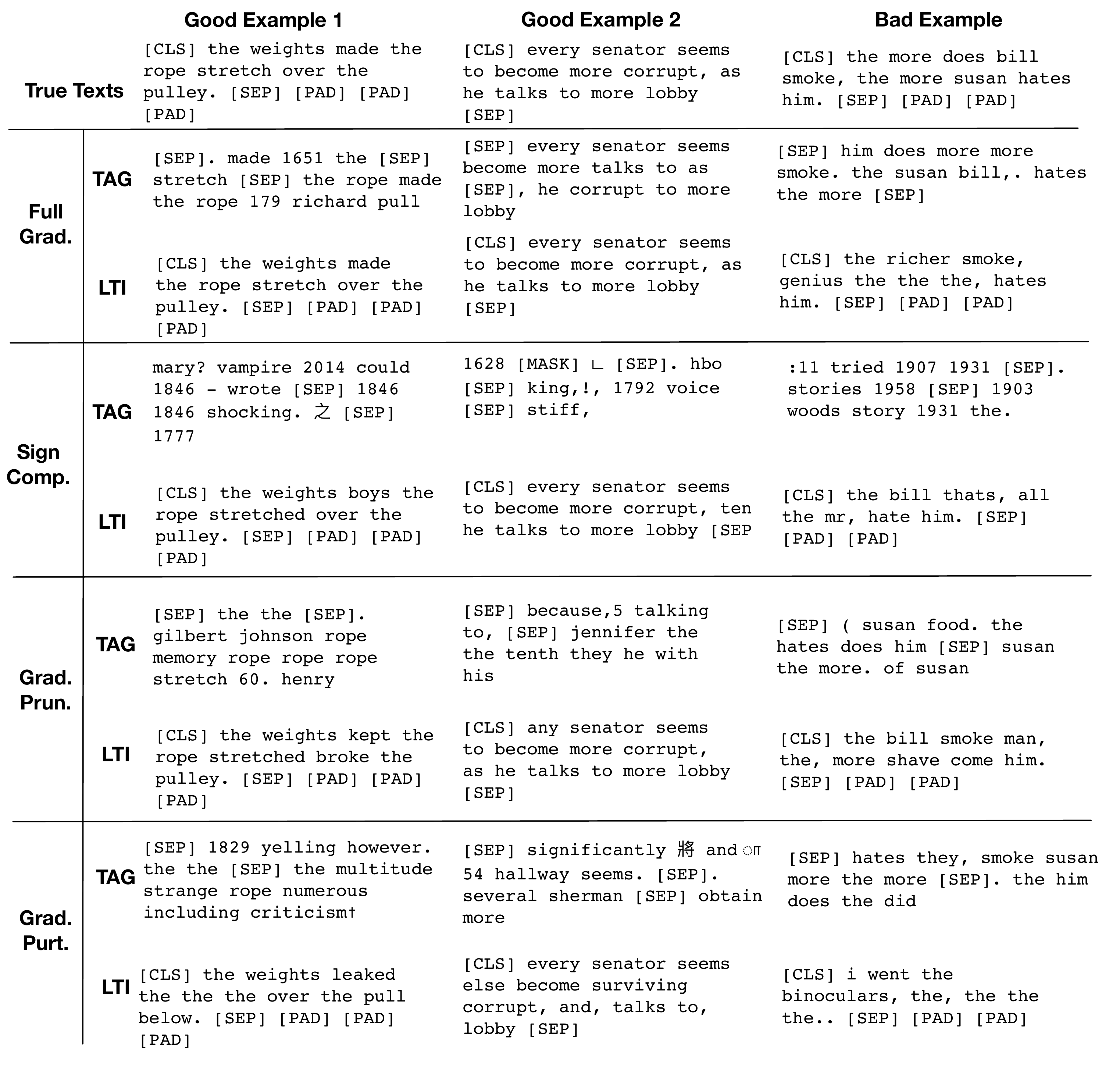}
    \caption{Samples from CoLA and their reconstructions.}
    \label{fig:text_example_cola}
\end{figure}

\begin{figure}[h]
    \centering
    \includegraphics[width=\linewidth]{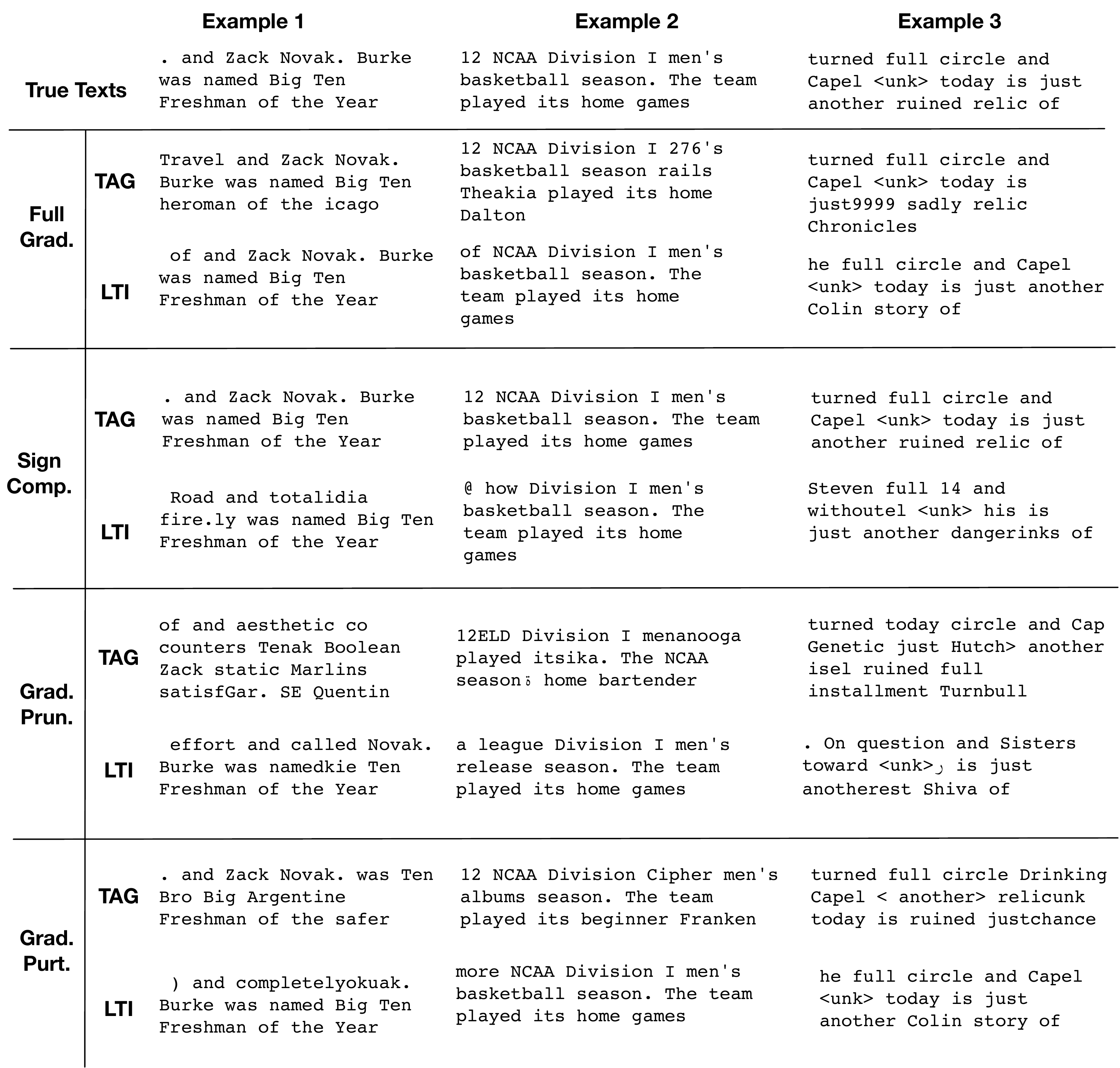}
    \caption{Samples from Wikitext and their reconstructions.}
    \label{fig:text_example_wikitext}
\end{figure}

\subsection{Examples on Vision Data}
\autoref{fig:img_example_more} shows additional samples and the reconstructions of attacks under various defense mechanisms on CIFAR10 dataset when the gradients are computed from LeNet. Similar to what we observe from the figure in the main text, all attacks can mostly reconstruct the data when there is no defense mechanism applied, while LTI is the only successful method when the defense mechanisms are applied. 

\autoref{fig:img_example_resnet} shows the examples when the FL model is ResNet20. We can observe that LTI is the only method that can reveal the partial object information of the original images across all gradient settings (including the setting where no defense mechanism is applied.)

\subsection{Examples on Language Data}
\autoref{fig:text_example_cola} shows three samples, including two good examples and one bad example (w.r.t. LTI), from CoLA dataset and their reconstructions when different defense mechanisms are applied.
The first observation is that LTI significantly performs better than TAG especially when the defense mechanisms are applied. Moreover, we find that the reconstruction error types of the two methods are different.
The error of TAG comes from both the wrong token prediction and the wrong token position prediction. In the reconstruction of TAG, many random tokens appear. 
Though the error of TAG is mostly the wrong token prediction, while the wrong tokens are the tokens with the high frequencies such as "the".

We also show three samples from WikiText dataset and the gradient inversion results from TAG and LTI in \autoref{fig:text_example_wikitext}. 
The comparison between TAG and LTI matches the results of quantitative evaluation in the main text: TAG has perfect performance when sign compression is applied, while LTI outperforms TAG in the other three settings.

\end{document}